\documentclass[letterpaper]{article} 

\usepackage{sty-TGVC}  
\usepackage{times}  
\usepackage{helvet}  
\usepackage{courier}  
\usepackage[hyphens]{url}  
\usepackage{graphicx} 
\usepackage{natbib}  
\usepackage{caption} 
\usepackage{algorithm}
\usepackage{algorithmic}

\usepackage{newfloat}
\usepackage{listings}

\usepackage{subfigure}
\usepackage{multirow}
\usepackage{tabularx}
\usepackage{amsmath}
\usepackage{booktabs}

\DeclareCaptionStyle{ruled}{labelfont=normalfont,labelsep=colon,strut=off} 
\floatstyle{ruled}
\newfloat{listing}{tb}{lst}{}
\floatname{listing}{Listing}
\nocopyright 

\title{T-GVC: Trajectory-Guided Generative Video Coding at Ultra-Low Bitrates}

\author {
    Zhitao Wang\textsuperscript{\rm 1},
    Hengyu Man\textsuperscript{\rm 1}\thanks{Corresponding author},
    Wenrui Li\textsuperscript{\rm 1},
    Xingtao Wang\textsuperscript{\rm 1},
    Xiaopeng Fan\textsuperscript{\rm 1,2,3},
    Debin Zhao\textsuperscript{\rm 1}
}
\affiliations {
     \textsuperscript{\rm 1}Harbin Institute of Technology\\ \textsuperscript{\rm 2}Harbin Institute of Technology Suzhou Research Institute\\ \textsuperscript{\rm 3}Peng Cheng Laboratory\\
    zhitao.wang.hit@outlook.com, manhengyu@hotmail.com, liwr618@163.com,\\\{xtwang, fxp, dbzhao\}@hit.edu.cn
}

\begin{document}

\maketitle

\begin{abstract}
Recent advances in video generation techniques have given rise to an emerging paradigm of generative video coding for Ultra-Low Bitrate (ULB) scenarios by leveraging powerful generative priors. However, most existing methods are limited by domain specificity (e.g., facial or human videos) or excessive dependence on high-level text guidance, which tend to inadequately capture fine-grained motion details, leading to unrealistic or incoherent reconstructions. To address these challenges, we propose \textbf{T}rajectory-Guided \textbf{G}enerative \textbf{V}ideo \textbf{C}oding (dubbed T-GVC), a novel framework that bridges low-level motion tracking with high-level semantic understanding. T-GVC features a semantic-aware sparse motion sampling pipeline that extracts pixel-wise motion as sparse trajectory points based on their semantic importance, significantly reducing the bitrate while preserving critical temporal semantic information. In addition, by integrating trajectory-aligned loss constraints into diffusion processes, we introduce a training-free guidance mechanism in latent space to ensure physically plausible motion patterns without sacrificing the inherent capabilities of generative models. Experimental results demonstrate that T-GVC outperforms both traditional and neural video codecs under ULB conditions. Furthermore, additional experiments confirm that our framework achieves more precise motion control than existing text-guided methods, paving the way for a novel direction of generative video coding guided by geometric motion modeling.
\end{abstract}


\section{Introduction}

One of the core challenges in video coding lies in effectively modeling inter-frame dependencies to reduce temporal redundancy and enhance coding efficiency. While traditional video coding standards (e.g., H.266/VVC \cite{bross2021overview} and AVS3 \cite{zhang2019recent}) have achieved remarkable progress, their reliance on handcrafted motion compensation and transform-quantization modules exhibits limitations in modeling non-rigid motions and nonlocal spatio-temporal dependencies, particularly under Ultra-Low Bitrate (ULB) scenarios with constrained bandwidth. Benefiting from the development of deep learning, recently emerging neural network-based video coding approaches \cite{10155660,li2024neural,jiang2024ecvc,Man2024,li2024object,Man2025,tang2025neural} have demonstrated promising potential to surpass conventional schemes. However, these methods primarily emphasize pixel-level signal fidelity and suffer from semantic information loss in ULB scenarios due to limited temporal context information. Although recent attempts \cite{yang2022perceptual,du2024cgvc,du2022generative} have leveraged GAN \cite{goodfellow2020generative} to improve the perceptual quality, they still adopt similar frameworks as \cite{li2023neural,li2024neural,yang2020rlvc} to model temporal contextual information, which constraints their ability to achieve lower bitrates. Consequently, preserving critical semantic information while enhancing perceptual quality has become a pressing challenge for ULB video compression.

Recent breakthroughs in video generation technology have provided new feasibility for video coding under ULB conditions, catalyzing the emergence of the concept of ``generative video coding" \cite{chen2024generative}. Unlike generation tasks in computer vision, generative video coding primarily aims to achieve content-faithful reconstruction under strict bitrate constraints by leveraging the strong priors of generative models as well as compact spatio-temporal guidance. Despite these theoretical advantages, most existing generative video coding schemes are tailored to specific types of videos, such as human face video \cite{chen2024generative2}, human body video \cite{wang2023extreme,chen2025rethinking}, and small motion video \cite{yin2024compressing}, due to the limited capabilities of earlier-stage generative models.

\begin{figure}
    \centering
    \includegraphics[width=0.95\linewidth]{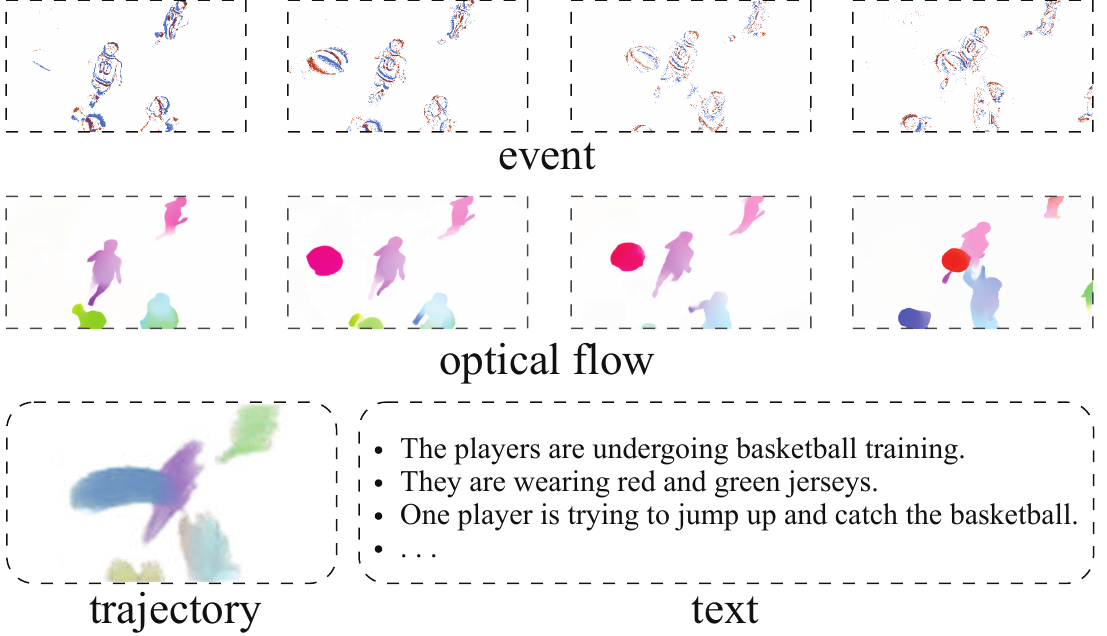}
    \caption{Examples of motion representation for video (event, optical flow, trajectory and text).}
    \label{fig1}
\end{figure}

The rapid advancement of video diffusion models (VDMs) \cite{xing2024survey} has significantly intensified research interest in extending generative video coding to natural video content. Specifically, some recent works \cite{zhang2024video,wan2025m3} attempt to model spatio-temporal information based on texts and keyframes, showing the potential to synthesize motion sequences conditioned on sparse and high-level information derived from the original video. Nevertheless, these methods still struggle to ensure temporal consistency and visual stability when applied to complex real-world scenarios. The generated outputs may exhibit unexpected artifacts, such as unnatural motion patterns or missing dynamic objects, compromising the temporal semantic fidelity with respect to the original video. To enable more precise natural video generation, an intuitive idea is to take advantage of low-level motion information, as illustrated in Figure \ref{fig1}. Unfortunately, employing dense motion representation incurs a higher bitrate overhead for VDMs to synthesize high-fidelity videos. \textbf{It is crucial to develop an effective strategy to extract compact motion guidance that preserves key temporal semantic information from the original video.}

Another important aspect concerns the conditional guidance mechanisms employed in the generative model. Most existing controllable video generation methods utilize `classifier-free guidance' \cite{ho2022classifier} to fine-tune pre-trained video diffusion models, such as Stable Video Diffusion (SVD) \cite{blattmann2023stable} and VideoCraft \cite{chen2023videocrafter1,chen2024videocrafter2}, or adopt a ControlNet-like adapter \cite{zhang2023adding,niu2024mofa,liu2025sketchvideo} to steer the generation process according to user specifications. However, the loss of critical structural and textural details in compressed keyframes severely impairs the model’s ability to reconstruct motion-aligned videos, particularly under ULB conditions. Moreover, these pre-trained generative models suffer from limited adaptability in the context of video coding tasks, as the compact temporal guidance (e.g., motion trajectory) may deviate substantially from the original conditioning domains. \textbf{How to effectively leverage compact guidance information to control the generative model in reconstructing high-fidelity and motion-aligned videos also requires further exploration.}

In response to the critical challenge of generative video coding under ULB conditions, we propose a \underline{T}rajectory-Guided \underline{G}enerative \underline{V}ideo \underline{C}oding framework. To balance the trade-off between temporal semantic information preservation and bitrate saving, a semantic-aware sparse motion sampling pipeline is first proposed to bridge low-level motion tracking with high-level semantic understanding. Specifically, we track the pixel-wise motion on a pre-defined grid and classify them into distinct motion instances according to motion patterns. To further improve coding efficiency, a sparse sampling of motion instances is performed based on their semantic importance, representing motion as a set of sparse trajectory points. The variations in trajectory points capture diverse motion patterns (such as translation, occlusion, deformation, etc.), which preserves critical temporal semantic information of the original video while significantly reducing the bitrate. To guide the diffusion-based generative model in reconstructing motion-aligned videos without compromising its inherent generative capabilities, we propose a training-free guidance approach. In contrast to existing methods that constrain on intermediate feature maps, our approach directly imposes constraints on the latent space of diffusion model via a lightweight yet effective guidance function, ensuring that the overall trajectories of target motion instances align with real-world motion paths.

To this end, our principal contributions are summarized as follows:

\begin{itemize}
\item We propose a semantic-aware sparse motion sampling pipeline tailored for generative video coding, which bridges low-level motion tracking with high-level semantic understanding by encoding motion as sparse trajectory points, significantly reducing bitrate while preserving temporal semantics.
 
\item A training-free latent space guidance mechanism is designed to enforce trajectory alignment via a lightweight guidance function without compromising the inherent capabilities of the generative model.

\item Extensive experiments demonstrate that T-GVC outperforms both traditional codecs and state-of-the-art deep learning-based methods under ULB conditions, establishing a novel paradigm for efficient semantic-aware video coding in resource-constrained scenarios.
\end{itemize}

\section{Related work}

\begin{figure*}[t!]
    \centering
    \includegraphics[width=1\linewidth]{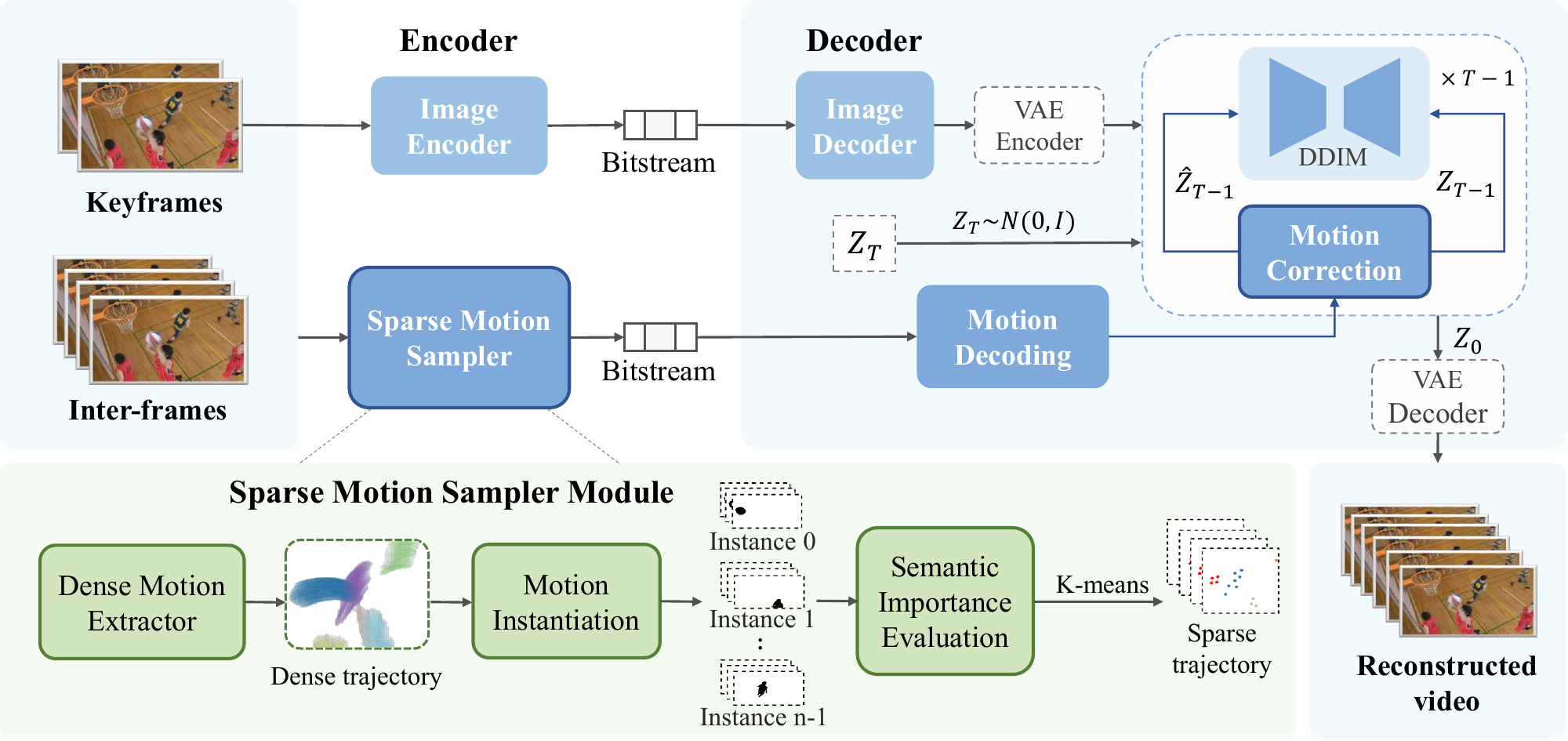}
    \caption{Overview of our T-GVC framework. On the encoder side, each pair of keyframes and corresponding inter-frames are fed into proposed sparse motion sampler to extract motion trajectories. Subsequently, the keyframes and trajectories are encoded into compact bitstreams. On the decoder side, each decoded keyframe pair is encoded into latent features via VAE encoder. These latent features, combined with zero-initialized latent features, form a latent sequence and concatenated with the initial latent noises as input of VDM. The decoded sparse motion trajectories act as guidance conditions during the inference process to correct the motion in latent space. Ultimately, the clean output is decoded by VAE. }
    \label{fig2}
\end{figure*}

\subsection{Generative video coding}

To ensure satisfactory perceptual quality, generative video coding has been explored in Ultra-Low Bitrate (ULB) scenarios. Early methods in this domain mainly focused on specific content types, such as facial videos \cite{chen2024generative2} or human-centric videos \cite{wang2022disentangled,wang2023extreme}. For natural videos, some approaches \cite{yang2022perceptual,du2022generative,du2024cgvc} employed GANs to enhance perceptual quality under low bitrate conditions. A more recent framework based on VQ-VAE \cite{qi2025generative} further performed transform coding in the latent space to achieve high realism and high fidelity at ULB. However, constrained by the inherent architectures of conventional frameworks \cite{yang2020rlvc,li2023neural,li2024neural}, these methods are difficult to generalize to even lower bitrate scenarios. 

Inspired by Cross-Modal Compression (CMC) in the field of image compression \cite{li2021cross,zhang2023rethinking,gao2023cross,gao2024rate}, \citeauthor{zhang2024video} designed Cross-Modal Video Compression (CMVC) to explore the potential of applying multimodal large language models for ULB video compression. In CMVC, a video is segmented into multiple semantically coherent clips, each of which is decomposed into content (a keyframe) and motion (text descriptions)\cite{10003241}. On the decoder side, keyframes are reconstructed first, followed by the generation of intermediate frames guided by the extracted `motion' information. Similarly, \citeauthor{wan2025m3} utilized a text-guided video diffusion model for video reconstruction, incorporating a more elaborate keyframe selection strategy. While these approaches are able to preserve essential semantic contents at ULB, they still suffer from issues with temporal consistency when handling complex motion patterns. Overall, despite significant progress in generative video coding for specific content types, extending these techniques to natural video, particularly in dynamic and motion-rich scenes, remains an open and challenging topic.

\subsection{Motion-conditioned diffusion model}
Text-conditioned video diffusion models (VDMs) have demonstrated remarkable capabilities in synthesizing visually striking content. However, they still struggle to achieve precise outcome control, frequently generating unrealistic sequences that violate physical plausibility. To improve temporal coherence and realism, recent efforts introduced motion guidance into VDMs. Early works like MCDiff \cite{chen2023motion} and DragNUWA \cite{yin2023dragnuwa} incorporated sparse motions and multimodal inputs, while later models such as MotionCtrl \cite{wang2024motionctrl} and Tora \cite{zhang2024tora,zhang2025tora2} introduced finer-grained control via explicit trajectory modeling. However, they typically require training new diffusion models or motion control modules from scratch. To mitigate the trade-off between quality and computation cost, more recent studies \cite{xiao2024video,zhang2025training} have explored the use of classifier guidance \cite{dhariwal2021diffusion} to control the video generation process in a more efficient manner, revealing the presence of intrinsic motion-aware features within the latent representations of VDMs. These approaches have shown the flexibility and effectiveness of the training-free video motion control framework, which provided valuable inspiration for our work.

\section{Methodology}
The overall structure of the proposed T-GVC framework is shown in Figure \ref{fig2}, which follows the common paradigm of generative video coding. In the remainder of this section, we will describe the entire coding pipeline of T-GVC and elaborate on each module with key symbols summarized in Table \ref{tab0}.

\subsection{Keyframe selection and compression}
Following \cite{zhang2024video,11146420}, we segment the source video into clips and select keyframes based on minimal pairwise semantic similarity, measured via a CLIP-based criteria \cite{radford2021learning}. To robustly handle scene transitions, we further incorporate a semantic-aware scene cut detection mechanism based on inter-frame similarity analysis. Specifically, when the semantic similarity between two consecutive frames falls below a pre-defined threshold, both frames are designated as keyframes to ensure seamless scene transitions. Therefore, more than one keyframes may be selected within each video clip. Each pair of consecutive keyframes $\{K_i, K_{i+1}\}$, along with the associated intermediate frames, constitutes a new clip $\{F_0,F_1,...,F_{L-1}\}$. The clip structure serves as the fundamental processing unit for subsequent motion sampling and semantic-consistent video reconstruction.


 Notably, since keyframes encapsulate critical semantic information, their reconstruction on the decoder side must maintain high perceptual quality even under stringent bitrate constraints, thereby preserving essential textural details and structural features to facilitate subsequent intermediate frame reconstruction. As this work primarily focuses on inter-frame coding, we adopt the off-the-shelf MS-ILLM \cite{muckley2023improving} for keyframe compression, owing to its demonstrated ability to preserve superior perceptual fidelity under aggressive bitrate constraints.

\begin{table}
  \setlength{\tabcolsep}{2.1mm}
  \centering
  \scalebox{1}{
\begin{tabular}{cc}
\toprule
    \textbf{Symbol}&\textbf{Description} \\
\midrule
$F_i$ & The $i$-th frame in the video clip. \\
$\mathcal{T}$ & Set of motion trajectories. \\
$\mathcal{T}^i$ & The $i$-th motion instance (cluster). \\
$T_i \in \mathcal{T}$ & The $i$-th trajectory in the trajectory set. \\
$(x_j^i, y_j^i)$ & Spatial coordinates of $T_i$ in $F_j$. \\
$V_j^i \in \{0,1\}$ & Visibility of $T_i$ at $F_j$. \\
$M_i$ & Motion mask of $\mathcal{T}^i$. \\
$z_t^{i}$ & The $i$-th latent feature at time step $t$. \\
$\tau_i^j$ & Sparse trajectory points of $\mathcal{T}^j$ in $F_i$. \\
$L$ & Number of frames in the video clip. \\
\bottomrule
\end{tabular}
}
\caption{Summary of symbols used in T-GVC.}
\label{tab0}
\end{table}

\subsection{Semantic-aware sparse motion sampling}
To bridge low-level motion tracking with high-level semantic understanding, we propose a semantic-aware sparse motion sampling framework that quantifies the semantic importance of different motion trajectories in the original video. The pipeline comprises three stages:
\paragraph{Dense trajectory extraction.}
Given two keyframes $\{K_i, K_{i+1}\}$ and their intermediate frames, we design a bidirectional tracking scheme to extract dense motion trajectories with improved temporal consistency. For each clip $\{F_0,F_1,...,F_{L-1}\}$ with temporal length $L$, we initialize isometric grids (grid size $=N_{grid}\times N_{grid}$) in the first frame and employ Co-Tracker \cite{karaev2024cotracker} to trace pixel-wise displacements originating from the grid vertices. The generated forward trajectory sequences over $L$ frames are represented as temporal chains of 3D coordinate points:
\begin{equation}
    \mathcal{T}_{fwd}=\left\{\left.T_i^{fwd}\right|T_i^{fwd}={(x_t^i,\ y_t^i,V_t^i)}_{t=0}^{L-1}\right\}
\end{equation}

To mitigate trajectory loss caused by occlusions or rotation, we reverse the temporal order of each clip and extract the corresponding backward trajectory sequences  $\mathcal{T}_{bwd}$ using the same procedure. Subsequently, the backward trajectories are temporally realigned and concatenated with their forward counterparts to synthesize the final dense trajectories, which holistically preserve the motion characteristics between keyframes:

\begin{equation}
    \mathcal{T}_{dense}={ \mathcal{T}}_{fwd} \cup Flip(\ \mathcal{T}_{bwd}\ )
\end{equation}
$Flip\left(\cdot\right)$ denotes the reverse operation in the time dimension.

\paragraph{Motion instantiation.}
To differentiate between distinct motion patterns, the raw trajectories are clustered using HDBSCAN \cite{mcinnes2017hdbscan}, a hierarchical density-based algorithm particularly effective in handling variable cluster densities and suppressing noise, which is critical for processing real-world videos with irregular motions. In this stage, $T_i$ is first transformed into 2D trajectories by excluding the points where $V_t^{i\ } = 0$. Then, we formulate each $T_i$ as a spatio-temporal feature vector according to their coordinates. By clustering these features, the corresponding trajectories are adaptively categorized into different motion instances $\mathcal{T}^0, \mathcal{T}^1,..., \mathcal{T}^{n-1}$ and form the corresponding motion masks, as illustrated in Figure \ref{fig2}. 


\paragraph{Semantic importance evaluation.}
The extracted dense motion trajectories inherently encapsulate the spatio-temporal dynamics of the source video, enabling the decoder to reconstruct inter-frame semantic relationships through diffusion-based generation (Section \ref{sec3.3}). However, noisy regions may be mistakenly identified as motion instances due to inaccuracies in motion clustering. Additionally, naively encoding these raw trajectories introduces significant bitrate overhead. To mitigate the influence of the accuracy of motion clustering and reduce inter-frame bitrate while preserving semantically critical motion information, we propose a trajectory sampling strategy driven by semantic loss awareness.

For each motion instance $\mathcal{T}^i$, we calculate the semantic importance score of motion $S^i_{inter}$ by multiplying the intra-semantic importance score $S^i_{intra}$ with the length of the trajectory. The intra-semantic importance score is defined as:
\begin{align}
S^i_{\text{intra}} 
&= \sum_{j=0}^{L-2} \big\| D(F_j, F_{j+1}) \notag \\
&- D\left[ P(F_j, M^i), P(F_{j+1}, M^i) \right] \big\|_1
\end{align}

where $D(\cdot)$ denotes the CLIP-based similarity, and $P(F_j, M^i)$ represents that the region covered by the motion mask $M^i$ in $F_j$ is replaced by the surrounding pixels to avoid feature mismatch for CLIP. 

We select the motion whose inter-frame semantic importance exceeds an empirically chosen threshold and extract sparse trajectories $\tau^i$ through k-means clustering, where the keypoint quantity $K$ for each instance is determined by the semantic importance score $S^i_{inter}$ and the quantity of trajectories $N^i$:
\begin{equation}
    K^i = min\{(\alpha\cdot S^i_{inter} + \beta\cdot \frac{N^i}{N_{\text{total}}} ) \cdot {K_{max}}, K_{max}\}
\end{equation}
$N_{total}$ represents the grid quantity while hyperparameters $\alpha$ and $\beta$ control the weighting ratios of different metrics.

Considering that the video resolution is downsampled by a factor of 8 after projected into the diffusion latent space on the decoder side, the trajectory coordinates are quantized into the same numerical range to ensure alignment with the compressed latent representation. Finally, the initial trajectory data from the first frame, along with the coordinate displacements between adjacent frames are losslessly encoded into a compact bitstream, which would be decoded to guide the reconstruction of the inter-frames on the decoder side.

\begin{figure}
\centering
\subfigure[Reconstructed frame 1]{
		\includegraphics[scale=0.25]{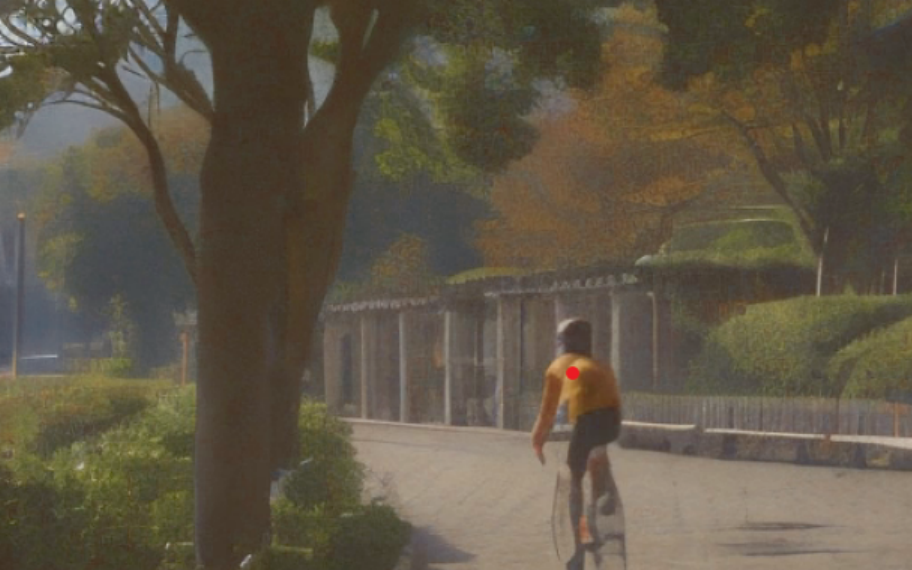}
        \label{fig3a}}
\subfigure[Latent feature map of frame 1 (T=9)]{
		\includegraphics[scale=0.25]{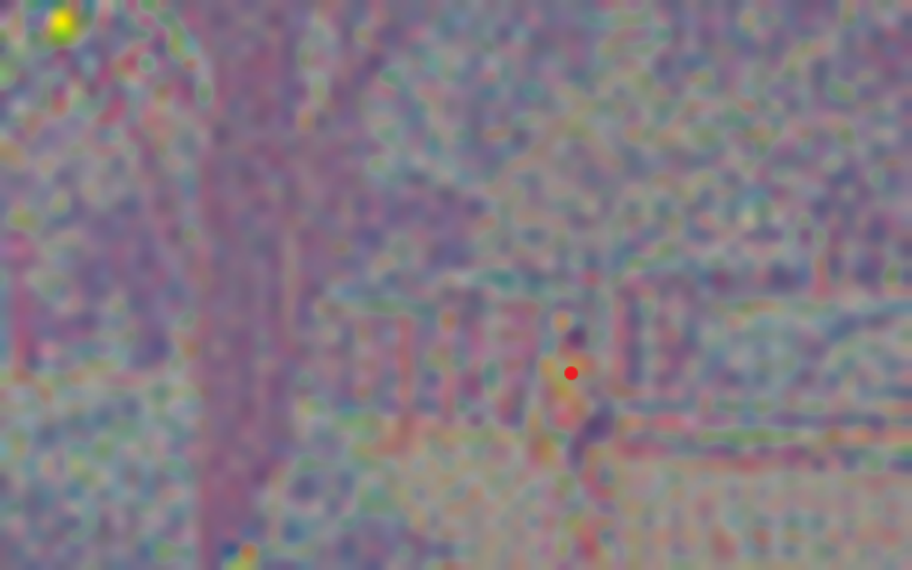}
        \label{fig3b}}
\\
\subfigure[Reconstructed frame 2]{
		\includegraphics[scale=0.25]{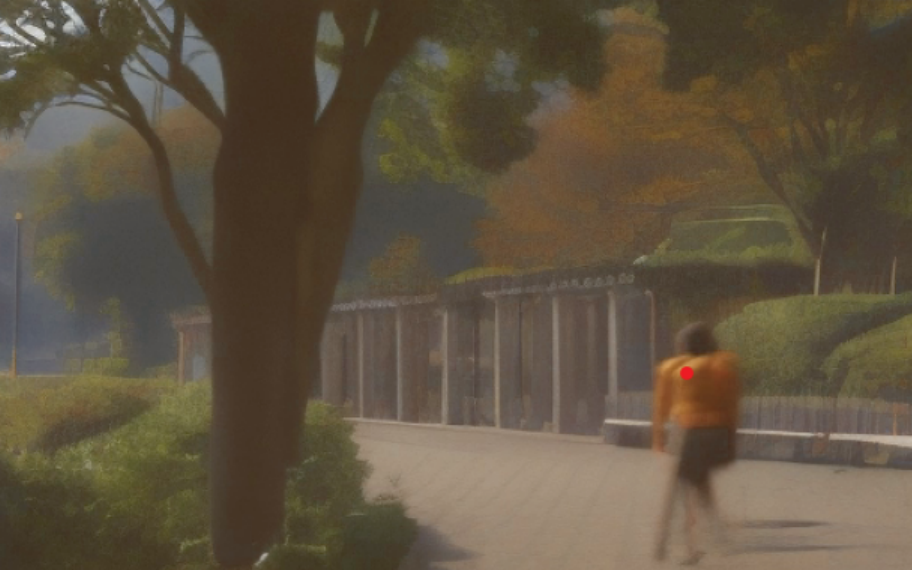}
        \label{fig3c}}
\subfigure[Heatmap on latent feature map of frame 2 (T=9)]{
		\includegraphics[scale=0.25]{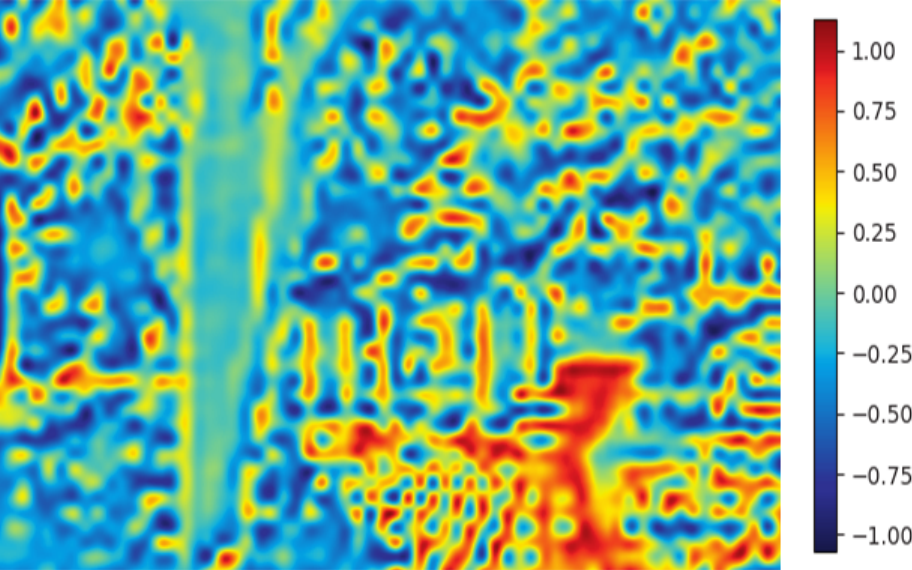}
        \label{fig3d}}
\caption{Illustration of the semantic correlation of latent features (upscaled to the same resolution as the original frame) in the same trajectory across two reconstructed frames. Given the red trajectory point in (b), we plot (d) according to the similarity between the feature on the point and the latent features of frame 2.}
\label{fig3}
\end{figure}

\subsection{Video reconstruction}
\label{sec3.3}
\paragraph{Trajectory-based motion guidance.}
Inspired by \cite{xiao2024video,zhang2025training}, we utilize motion trajectories to guide the generation process in a training-free manner by imposing trajectory-aware constraints during the denoising diffusion process. In \cite{xiao2024video,zhang2025training}, the trajectories are extracted from the feature maps of U-Net blocks in VDM. However, these approaches incur substantial computational overhead during inference,  particularly for generative models with a large number of parameters. Moreover, since the motion of generated results is globally controlled, it is difficult to guide the model's output with fine-grained precision. In contrast, T-GVC directly imposes constraint on the latent noise, leveraging the insight that the latent space in video diffusion contains crucial structural information and semantic correlation. As shown in Figure \ref{fig3}, the region with the highest similarity in Figure \ref{fig3a} corresponds to the position of the red point in Figure \ref{fig3c}, which belongs to the same trajectory as the red point in Figure \ref{fig3a} or Figure \ref{fig3b}.

For each clip of the target video with length $L$, the pipeline on the decoder side begins with decoding compressed keyframes and associated motion trajectories. After that, the keyframes $\{{\hat{K}}_i, {\hat{K}}_{i+1}\}$ are projected into latent space via VAE \cite{kingma2013auto} encoder and concatenated with the initial noisy latent $z_T= \{z_T^0,\ z_T^1,\ \ldots,\ z_T^{L-1}\}$ as conditions, providing the key textual and structural features of the original video. 

Let $F\left(z_t^i,\tau_i^j\right)$ denote the feature of the $i-th$ latent $z_t^i$ at time step t within the region covered by the sparse trajectories $\tau_i^j$ of the motion instance $\mathcal{T}^j$. To reconstruct the temporal semantic information in the guidance of trajectories, we define a loss function to measure how well the trajectories of $z_t$ align with the trajectory guidance, thereby constraining the denoising process:
\begin{equation}
    \mathcal{L}_m=\ \sum_{i=1}^{L-2}\sum_{j=0}^{n-1}\left[\alpha_i\cdot  S_0\left(z_t^i,\tau_i^j\right)+\beta_i\cdot S_{L-1}\left(z_t^i,\tau_i^j\right)\right]
\end{equation}
where $S_k(z_t^i,\tau_i^j)=\|{F(z_t^k,\tau_k^j)- F(z_t^i,\tau_i^j)}\|_1$, captures the non-local similarity of the latent sequence, while $\alpha_i$ and $\beta_i$ control the level of the similarity.

\begin{figure*}[!t]
    \centering
    \includegraphics[width=0.95\linewidth]{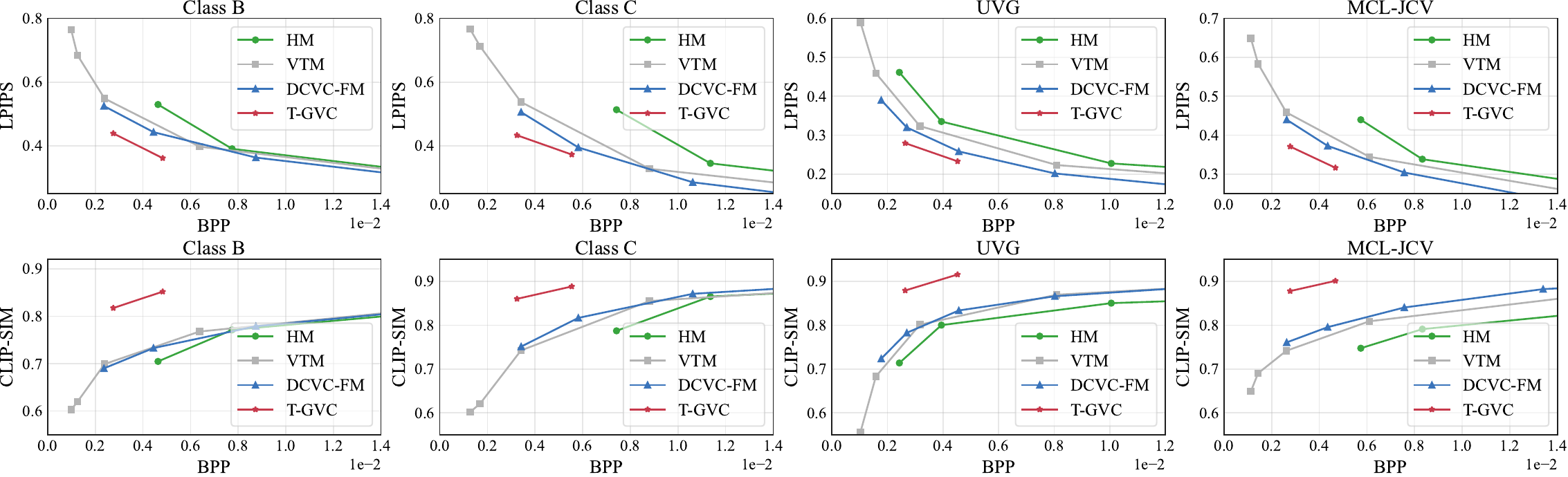}
    \caption{The R-D performance comparison results for HEVC Class B, Class C, UVG and MCL-JCV datasets.}
    \label{fig5}
\end{figure*}

We can further update the output $\epsilon_\theta$ of the model in each time step via the gradient of $\mathcal{L}_m$:
\begin{equation}
{\hat{\epsilon}}_\theta{\left(z_t,t\right)}=\epsilon_\theta{\left(z_t,t\right)}+{s\left(t\right)}\cdot\nabla _{z_t}\mathcal{L}_m\left({f(z_t)}\right)
\end{equation}
where $f(\cdot)$ is the map from noise latent $z_t$ to the clean latent $z_0$, and a detailed theoretical explanation of our proposed guidance mechanism is detailed in the appendix.


\begin{figure}
    \centering
    \includegraphics[width=0.95\linewidth]{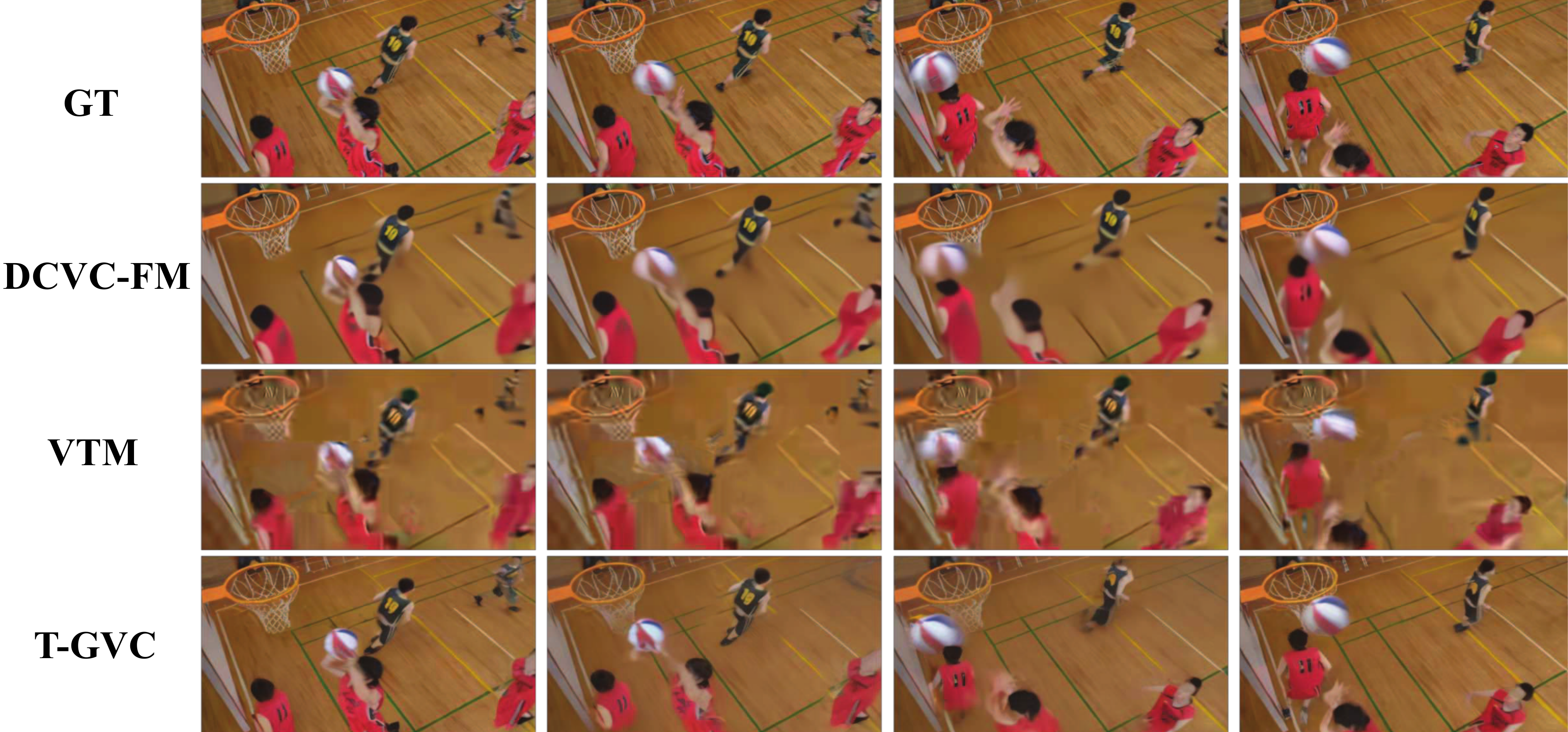}
    \caption{Visual quality comparison: ground truth, DCVC-FM, VTM and proposed T-GVC (top to bottom). The reconstructed frames of our framework demonstrates higher perceptual quality at similar bitrates.}
    \label{fig6}
\end{figure}

\paragraph{Video generation with variable length.}
Most existing VDMs are trained to generate videos of fixed length (e.g., 16 frames). To this end, we further design a variable-length generation scheme to enhance the flexibility of pre-trained VDMs. Specifically, for each video clip with length $L \leq 16$,  we first interpolate the sparse trajectories $\tau$ to $\hat{\tau}$ with length $\hat{L}=16$ and assign corresponding positional markers to the original trajectories. The denoised latent frames at these marked positions are then selected and concatenated to form the final video. For each video clip with length $L>16$, we perform a dual-stage generation strategy where the trajectories are divided into two segments, with each segment guiding one generation stage separately. 



\section{Experiments}
\subsection{Experimental setup}
\paragraph{Dataset and metrics.}
Following \cite{wan2025m3}, we select HEVC Class B, C \cite{bossen2010common} as well as UVG \cite{mercat2020uvg} and MCL-JCV \cite{wang2016mcl} as the test dataset to evaluate the rate-distortion (R-D) performance. All test videos are resized to $512 \times 320$ and consist of 96 frames with the original frame rate. For the calculation of R-D metric, the compression rate is quantified in bits per pixel (bpp), while the distortion is measured using LPIPS \cite{zhang2018unreasonable} for perceptual quality and CLIP-SIM \cite{radford2021learning} for semantic similarity. Lower LPIPS and higher CLIP-SIM at the same bitrate represent better coding performance.

\paragraph{Implementation details.}
We select the pre-trained DynamiCrafter \cite{xing2024dynamicrafter} video diffusion model as our inter-frame decoder due to its superior quality for open-domain image animation. We remove text prompts from the original model and solely utilize trajectories and keyframes to guide the generation of inter-frames. The number of DDIM denoising steps for each video and the classifier guide scale $s\left(t\right)$ are set to 10 and $30\cdot\sqrt{1-\alpha_t}$, respectively. For keyframe compression, we adopt MS-ILLM \cite{muckley2023improving} under `quality 1' and `quality 2' settings. In addition, given that the video resolution is downsampled to $64 \times 40$ by a factor of 8 after projected into latent space, the grid size for sparse motion sampling is configured as 64 to preserve structural consistency. All experiments are conducted on a single NVIDIA RTX 3090 GPU. 

\subsection{Quantitative and qualitative results}
We compare T-GVC with traditional codecs (H.265 \cite{sullivan2012overview} and H.266 \cite{bross2021overview}) and one of the most competitive open-source neural video codec (DCVC-FM \cite{li2024neural}) with a wide bitrate range to evaluate its coding performance at ULB. The comparison results with other generative video coding schemes are provided and discussed in Section \ref{sec4.3}. For H.265, we use HM-18.0 with QP = 51, 45 and 39. For H.266, we use VTM-23.7 with QP = 63, 57, 51, 45 and 39. For DCVC-FM, we set the $q$ indexes to 0, 8, 16, 24 and 32, which control the compression level. All codecs above are tested under Low Delay P configuration with intra-period$=$-1. 

The R-D curves for the test video classes are presented in Figure \ref{fig5}, where the bitrate of our T-GVC is controlled by adjusting the quality of keyframes. As observed, T-GVC outperforms both traditional codecs and the neural video codec in terms of perceptual reconstruction quality and semantic fidelity at ULB conditions. Moreover, benefiting from powerful generative priors, T-GVC achieves lower bitrate points (below 0.005 bpp) than GAN- or VQ-VAE-based methods \cite{yang2022perceptual,qi2025generative}. Figure \ref{fig6} shows the visual results for the sequence `BasketballDrill\_832x480\_50' from the HEVC Class C dataset. To ensure a fair comparison, reconstructed sequences from different methods are selected at comparable bitrates. It is observed that the reconstructed videos generated by T-GVC demonstrate superior fidelity in the background regions and key foreground objects (e.g., basketball and the players in red). In contrast, VTM exhibits noticeable blocking artifacts while DCVC-FM suffers from significant high-frequency detail loss, leading to degraded texture quality. 

It is worth noting that although some texture and structural details in the reconstructed frames might be partially missing or inaccurate to some extent (e.g., the direction of the ball and the appearance of players in green), the overall visual quality remains satisfactory for the ULB scenarios. Under ULB conditions, we primarily focus on perceptual quality within a given bitrate range. While key semantic information (e.g., the player jumping to catch the ball) is preserved, certain visual artifacts—such as the deblurring effects or minor spatial deviations—are acceptable.

\begin{table}
\centering
  \setlength{\tabcolsep}{2.3mm}
  \begin{tabular}{ccc}
    \toprule
    Models&Settings&BD-rate (\%)\\
    \midrule
    \multirow{2}{*}{Text} & W/o motion bitrate & -4.45\\
    & W/ motion bitrate & 31.92\\
    \midrule
    \multirow{2}{*}{\textbf{Trajectory}} & W/o motion bitrate & -16.84\\
    & W/ motion bitrate & \textbf{-3.06}\\
    \midrule
    \multirow{2}{*}{Text+Trajectory} & W/o motion bitrate & -13.33\\
    & W/ motion bitrate & 34.97\\ 
  \bottomrule
\end{tabular}
\caption{Ablation study on the guidance mechanism.}
  \label{tab1}
\end{table}

\subsection{Ablation study}
\label{sec4.3}
To further verify the rationality of our design, we conduct ablation studies on two core components of the T-GVC framework: the guidance mechanism and the motion sampler module. Video sequences characterized by large motions, including `ParkScene', `BasketballDrill', `PartyScene', `videoSRC02' and `videoSRC05' from Class B, Class C and MCL-JCV, are selected for evaluation. The coding efficiency is assessed using the Bjøntegaard Delta rate (BD-rate) \cite{bjontegaard2001calculation}, where PSNR is replaced by 1 - LPIPS. Negative BD-rate values indicate bitrate saving for equivalent perceptual quality. We apply DynamiCrafter with empty prompts (i.e., only guided by keyframes) as the comparison anchor in our experiments. The reported BD-rate results are calculated by averaging all test sequences.

\begin{figure}
    \centering
    \includegraphics[width=0.95\linewidth]{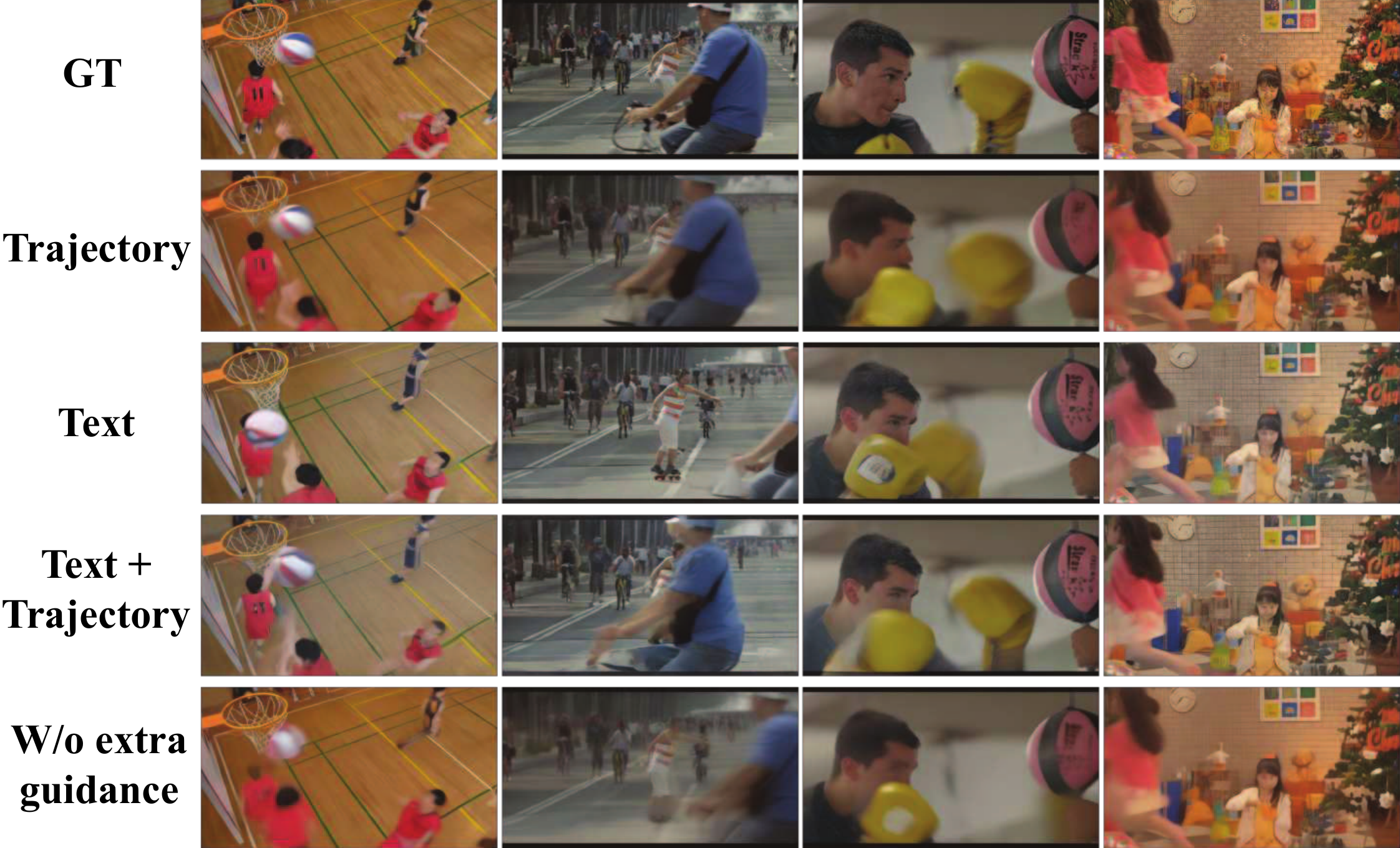}
    \caption{Visual quality comparison for ablation study on the guidance mechanism.}
    \label{fig7}
\end{figure}

\paragraph{Ablation study on the guidance mechanism.}
In prior generative video coding methods \cite{zhang2024video,wan2025m3}, text is commonly adopted as motion guidance. To demonstrate the superiority of our guidance mechanism, we implement three variants of generative frameworks based on DynamiCrafter: one guided by text (denoted as Text), one guided by trajectories (denoted as Trajectory) and the other by both text and trajectories (denoted as Text+Trajectory). Trajectories are sampled through our proposed sparse motion sampler, while text descriptions are generated from the source video using Hunyuan-Large \cite{sun2024hunyuan} without limitation on length. The comparison results are presented in Table \ref{tab1}, where "W/o motion bitrate" refer to excluding the bitrate consumed by motion-related information when calculating BPP, while "W/ motion bitrate" indicate that this bitrate is included. It is observed that the trajectory-guided approach consistently achieves better coding performance under both evaluation conditions. In particular, the perceptual quality improvement brought by trajectory guidance makes up for the additional bitrate overhead, offering a distinct advantage over text-based guidance. When overhead is not considered, the trajectory-guided method still outperforms the text-guided method. Combining text and trajectory guidance may enhance texture on simple-motion sequences. Some visual examples in Figure \ref{fig7} further demonstrate that trajectory guidance can perform more precise motion control, especially in dynamic areas. Text guidance can improve perceptual quality to some extent and combining text and trajectory guidance may enhance texture on simple-motion sequences. However, in more complex cases, text prompts may introduce artifacts that conflict with trajectory-aligned motion, leading to degraded results. For instance, as shown in the first column of Figure \ref{fig7}, the basketball appears behind the net, which violates the physical plausibility.

\begin{table}
\centering
  \setlength{\tabcolsep}{4.3mm}
\begin{tabular}{ccc}
\toprule
Models & Settings & BD-rate (\%) \\ 
\midrule
\multirow{2}{*}{Dense} & W/o motion bitrate & {-17.88} \\ 
 & W/ motion bitrate & {188.77} \\ \midrule
\multirow{2}{*}{Random} & W/o motion bitrate & {-7.95} \\ 
 & W/ motion bitrate & {15.05} \\ \midrule
\multirow{2}{*}{\textbf{Sparse}} & W/o motion bitrate & {-16.84} \\ 
 &  W/ motion bitrate & {\textbf{-3.06}}  \\ 
\bottomrule
\end{tabular}
\caption{Ablation study on the sparse motion sampler.}
  \label{tab2}
\end{table}

\paragraph{Ablation study on the motion sampler module.}
To validate the effectiveness of the semantic-aware motion sampling pipeline proposed in T-GVC, we further conduct an ablation study on the motion sampler module. The ablation results are presented in Table \ref{tab2}. Specifically, ``Sparse" refers to motion trajectories extracted with our semantic-aware motion sampling pipeline. ``Dense" denotes applying dense trajectories as guidance without sampling. ``Random" corresponds to randomly selecting sparse trajectories with $k$ keypoints ($K < K_{max}$) for each motion instance. We set the maximum keypoint quantity $K_{max}$ as 15 for the optimal R-D performance. Although dense trajectories can provide more precise motion representation for video reconstruction, such excessive bitrate overhead is intolerable for the video coding task. In contrast, our motion sampling pipeline achieves significant bitrate reduction in motion modeling, while preserving perceptual quality comparable to that of dense trajectory guidance. Compared with a random sampling strategy, our approach thoroughly considers the impact of each motion instance on the video semantics, effectively reducing the bitrate while retaining key motion information.

\section{Conclusion}
In this paper, we propose T-GVC, a novel Trajectory-Guided Generative Video Coding framework designed for ULB scenarios. Unlike existing methods that rely on domain-specific priors or high-level text guidance, T-GVC introduces a semantic-aware sparse motion sampling strategy, extracting trajectories based on semantic importance to retain key temporal information with minimal bitrate overhead. To further enhance motion realism, a training-free guidance mechanism is integrated into the diffusion process, enabling physically plausible reconstructions. Experimental results demonstrate that T-GVC significantly outperforms traditional codecs and state-of-the-art generative methods, especially in terms of motion control precision. 

\section{Acknowledgments}
This work was supported in part by the National Key R\&D Program of China (2023YFA1008500), the National Natural Science Foundation of China (NSFC) under grants U22B2035 and 62502116, and China Post-Doctoral Science Foundation under Grant 2025M774315.

\bibliography{bib-TGVC}

\appendix
\section{Theory}

\paragraph{Classifier guidance.}
Classifier guidance \cite{dhariwal2021diffusion} is a widely-used technique that modifies the sampling process of a pre-trained diffusion model. The primary objective of conditional generative models is to synthesize a target content $z$ given a conditioning signal $c$, ensuring that $z$ aligns with $c$. To formalize the conditional generation process, the target probability distribution $p\left(\left.z\right|c \right)$ can be derived through Bayes formula:
\begin{equation}
p\left(\left.z\right|c \right)=\frac{p \left(\left.c\right|z \right)p\left(z\right)}{p\left(c\right)}
\label{eq1}
\end{equation}
where $p\left(z\right)$ is unconditional probability distribution and $p \left(\left.c\right|z \right)$ is the probability distribution of the condition $c$ given a certain content $z$, while the prior probability $p\left(c\right)$ is independent of $z$.

And taking the gradient of the log likelihood of both sides of the Equation \ref{eq1}:
\begin{equation}
\mathrm{\nabla}_zlog{\left(p\left(\left.z\right|c\right)\right)}=\mathrm{\nabla}_zlog{\left(p\left(z\right)\right)}+\mathrm{\nabla}_zlog{\left(p\left(\left.c\right|z\right)\right)}
\label{eq2}
\end{equation}

Note that the denoising process in diffusion has a relationship with the gradient which represents score in Stochastic Differential Equation (SDE) \cite{song2020score}:
\begin{equation}
\epsilon_\theta\left(z_t,t\right)=-\sqrt{1-\alpha_t}\mathrm{\nabla}_{z_t}\log{\left(p\left(z_t\right)\right)}
\end{equation}
\begin{equation}
\epsilon_\theta\left(z_t,t,c\right)=-\sqrt{1-\alpha_t}\mathrm{\nabla}_{z_t}\log{\left(p\left(\left.z_t\right|c\right)\right)}
\end{equation}

We could further rewrite Equation \ref{eq2} as classification-guided sampling:
\begin{equation}
\epsilon_\theta\left(z_t,t,c\right)=\epsilon_\theta\left(z_t,t\right)-\sqrt{1-\alpha_t}\mathrm{\nabla}_{z_t}\log{\left(p\left({\left.c\right|z}_t\right)\right)}
\label{eq4}
\end{equation}
where $\sqrt{1-\alpha_t}\mathrm{\nabla}_{z_t}\log{\left(p\left({\left.c\right|z}_t\right)\right)}$ guides the sampling process.

Bansal \emph{et al.} \cite{bansal2023universal} extended Equation \ref{eq4} to universal guidance as
\begin{equation}
{\hat{\epsilon}}_\theta\left(z_t,t\right)=\epsilon_\theta\left(z_t,t\right)+s\left(t\right)\cdot{\nabla_{z_t}\ell\left(c,f(z_t)\right)}
\label{eq5}
\end{equation}
where $s\left(t\right)$ controls the guidance strength and the off-the-shelf guidance function $f\left( \cdot \right)$ and loss function $\ell\left( \cdot \right)$ could be calculated based on the predicted clean latent $z_0$. This theoretical foundation makes training-free motion guidance possible and inspires us to further explore a new paradigm in trajectory-based guidance.

\section{Technical detail}

\paragraph{Motion instantiation.}
To differentiate between distinct motion patterns, the raw trajectories are clustered using HDBSCAN \cite{mcinnes2017hdbscan}, a hierarchical density-based algorithm particularly effective in handling variable cluster densities and suppressing noise, which is critical for processing real-world videos with irregular motions. In this stage, $T_i$ is first transformed into 2D trajectories by excluding points where $V_t^{i\ } = 0$. Then, we formulate each $T_i$ as a spatio-temporal feature vector $f^i_{traj}$ as follows:
\begin{equation}
    f^i_{traj}=\left[x^i_0,\ y^i_0,\ \Delta x^i,\ \Delta y^i,\ d^i,\ \bar{\Delta\theta^i}\right]
\end{equation}
where $x^i_0$ and $y^i_0$ denote the initial coordinates in the reference keyframe, $\Delta x^i$ and$ \Delta y^i$ are the displacement components, $d^i$ is the total distance of the motion, and $\bar{\Delta\theta^i}$ represents the mean directional change calculated via the gradients of the coordinates. 

By clustering these features, the corresponding trajectories are categorized into different motion instances $\mathcal{T}^0, \mathcal{T}^1,..., \mathcal{T}^{n-1}$ and form the corresponding motion masks.

\begin{figure}
    \centering
    \includegraphics[width=1\linewidth]{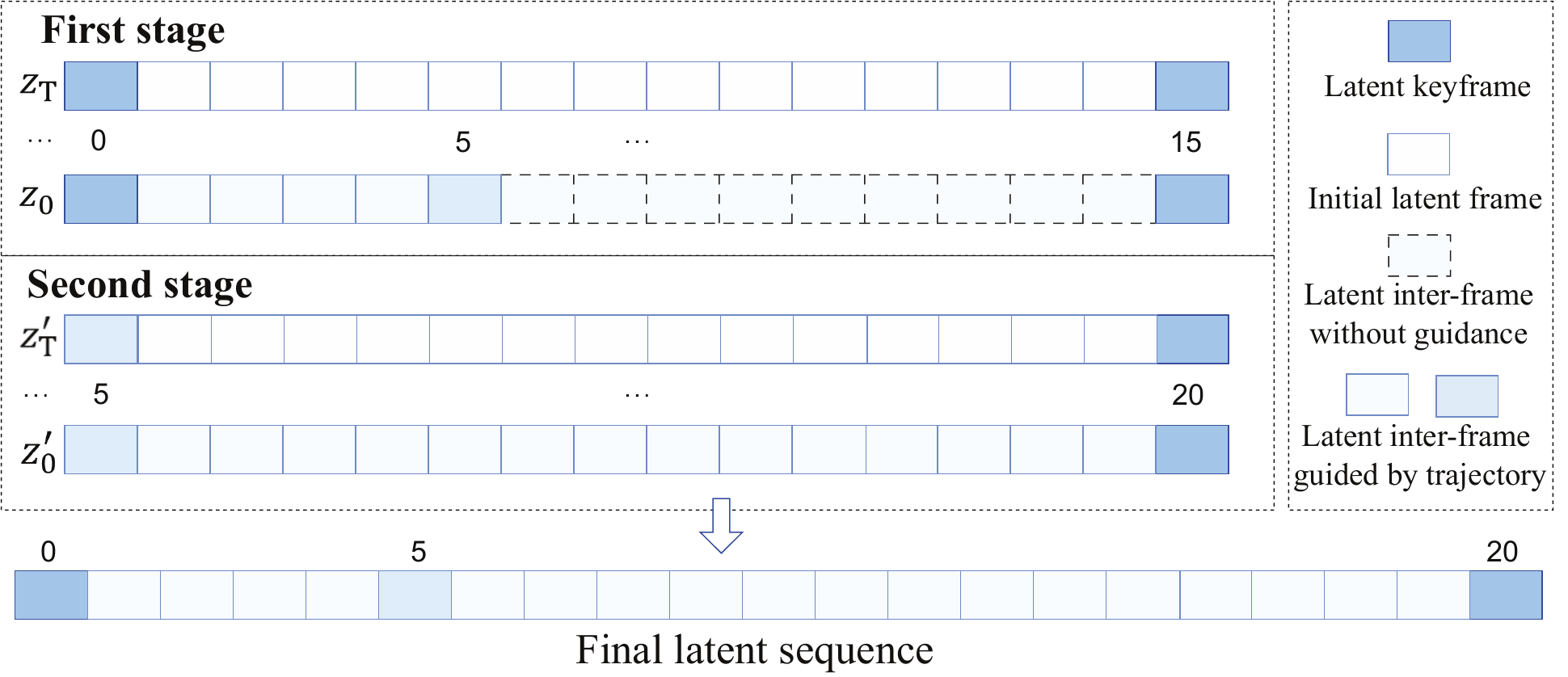}
    \caption{The generation process of our framework when the target video clip length exceeds 16 ($\mathbf{\boldsymbol{{L}={21}}}$).}
    \label{fig4}
\end{figure}

\begin{table}
\centering
  \renewcommand\arraystretch{1.1}
    \setlength{\tabcolsep}{0.9mm}
  \scalebox{1}{
\begin{tabular}{@{}clcl@{}}
\toprule
\multicolumn{2}{c}{Models} & Settings & \multicolumn{1}{c}{BD-rate (\%)} \\ \midrule
\multicolumn{1}{c}{\multirow{8}{*}{Sparse}} & \multirow{2}{*}{$K_{max}$ = 5} & W/o trajectory bitrate & \multicolumn{1}{c}{-5.94} \\ 
\multicolumn{1}{c}{} &  & W/ trajectory bitrate & \multicolumn{1}{c}{4.12}  \\ \cmidrule(l){2-4} 
\multicolumn{1}{c}{} & \multirow{2}{*}{$K_{max}$ = 10} & W/o trajectory bitrate & \multicolumn{1}{c}{-13.84}  \\ 
\multicolumn{1}{c}{} &  & W/ trajectory bitrate & \multicolumn{1}{c}{-1.86}  \\ \cmidrule(l){2-4} 
\multicolumn{1}{c}{} & \multirow{2}{*}{\textbf{$\mathbf{\boldsymbol{K_{max}}}$ = 15}} & W/o trajectory bitrate & \multicolumn{1}{c}{-16.84} \\ 
\multicolumn{1}{c}{} &  & W/ trajectory bitrate & \multicolumn{1}{c}{\textbf{-3.06}}  \\ \cmidrule(l){2-4} 
\multicolumn{1}{c}{} & \multirow{2}{*}{$K_{max}$ = 20} & W/o trajectory bitrate & \multicolumn{1}{c}{-10.58}  \\ 
\multicolumn{1}{c}{} &  & W/ trajectory bitrate & \multicolumn{1}{c}{5.82} \\
 \bottomrule
\end{tabular}
}
  \caption{Ablation study on the keypoint quantity.}
    \label{tab3}
\centering
\end{table}

\paragraph{Video generation with variable length.}
Figure \ref{fig4} demonstrates the operational scenario of our variable-length generation scheme in processing a video clip beyond 16 frames, with a theoretical maximum capacity of 30 frames. Specifically, for each video clip with length $L>16$, we perform a dual-stage generation strategy where the trajectories are divided into two segments, with each segment guiding one generation stage separately. Notably, increasing the number of processing stages facilitates extended video generation with enhanced temporal coherence, while linearly escalating computational demands.

\section{Ablation study}

\paragraph{Ablation study on the keypoint quantity.}
In the design of T-GVC, $K_{max}$ is a critical term that controls the maximum number of clustering central points of each motion instance. To evaluate its impact on coding performance, we perform ablation studies on $K_{max}$ by setting its value to 5, 10, 15, and 20, respectively. The comparison results are shown in Table \ref{tab3}. As observed, selecting an appropriate number of keypoints is essential for maintaining high-fidelity video reconstruction quality. Theoretically, under unconstrained bitrate conditions, increasing the keypoint quantity would lead to improved reconstruction quality. However, due to the limitations of current pixel-tracking models in terms of accuracy and stability, our method demonstrates a paradoxical phenomenon: excessive keypoints may introduce feature mismatch cascades that impair the inference process of the generative model, ultimately degrading reconstruction fidelity. 

\paragraph{Ablation study on the video reconstruction.}
In our study, DynamiCrafter \cite{xing2024dynamicrafter} was selected as the base model due to its superior reconstruction quality compared to other interpolation methods assessed in \cite{zhang2024video}. We conduct additional experiments by integrating diffusion-based SEINE \cite{chen2023seine} and flow-based Ltx-Video \cite{hacohen2025ltx} into T-GVC framework as base video models.

Comparison results at two different bitrate points are provided in Table \ref{tab5}. As observed, DynamiCrafter outperforms SEINE and Ltx-Video in terms of LPIPS. Notably, since Ltx-Video applies higher-magnitude compression in the spatiotemporal dimensions, it introduces significant positional deviation and reduce the precision of trajectory-to-feature alignment, leading to notable performance degradation. When using compressed keyframes and text as guidance, we observe that Ltx-Video would generate artifacts and unrealistic motion in the reconstruction, which is a common issue shared by other existing generative video models. 

\begin{table}
\centering
  \renewcommand\arraystretch{1.1}
  \setlength{\tabcolsep}{3.7mm}
\begin{tabular}{cccc}
\toprule
Models & LPIPS1 $\downarrow$ & LPIPS2 $\downarrow$ \\ 
\midrule
\multirow{1}{*}{SEINE} & \multirow{1}{*}{0.5201} & {0.5467} \\ 
\multirow{1}{*}{Ltx-Video} & \multirow{1}{*}{0.4959}  & {0.5201} \\  
\multirow{1}{*}{DynamiCrafter} & \multirow{1}{*}{0.3631} & {0.4123} \\ 
\bottomrule
\end{tabular}
\caption{Ablation study on the video reconstruction.}
  \label{tab5}
\end{table}

\begin{table}
\centering
  \renewcommand\arraystretch{1.1}
  \setlength{\tabcolsep}{0.6mm}
\begin{tabular}{cccc}
\toprule
Models & Interval & Settings & BD-rate (\%) \\ 
\midrule
\multirow{2}{*}{Fixed Interval} & \multirow{2}{*}{16} & W/o motion bitrate & {-30.81} \\ 
 &  & W/ motion bitrate & {21.15} \\ \midrule
\multirow{2}{*}{Fixed Interval} & \multirow{2}{*}{24} & W/o motion bitrate & {28.06} \\ 
 & \multirow{1}{*}{}& W/ motion bitrate & {43.69} \\ \midrule
\multirow{2}{*}{\textbf{Ours}} & \multirow{2}{*}{Adaptive} & W/o motion bitrate & {-16.84} \\ 
 & \multirow{1}{*}{}& W/ motion bitrate & {\textbf{-3.06}} \\
\bottomrule
\end{tabular}
\caption{Ablation study on the keyframe selection.}
  \label{tab4}
\end{table}

\paragraph{Ablation study on the keyframe selection.}
Although this adaptive approach is not the focus of our work and  has been validated in prior work \cite{zhang2024video}, we include additional ablation study using unguided adaptive keyframe selection as a baseline. As observed, a larger frame interval greatly degrade quality, while smaller one leads to higher reconstruction quality but results in increased overhead. In contrast, our adaptive method effectively balances these two aspects. 

\section{Limitation}

Despite impressive performance under ULB conditions, T-GVC still presents several limitations. In scenes involving occlusion or deformable motion, T-GVC captures motion modes via key point variation. However, in cases involving transformations, such as rotation, texture artifacts may emerge despite the preservation of 2D motion trajectories. For human-centric videos, although T-GVC generally preserves semantic fidelity, occasional visual artifacts such as blurriness or unnatural limb motion might occur due to the higher motion complexity (e.g., the reconstructed sequence ``videoSRC23\_1920x1080\_24" in the supplementary materials). Besides, the high computational demand of dense trajectory tracking and video reconstruction makes real-time application of T-GVC still a challenging task, which is a common challenge for generative codecs.

\begin{figure}
    \centering
    \includegraphics[width=1\linewidth]{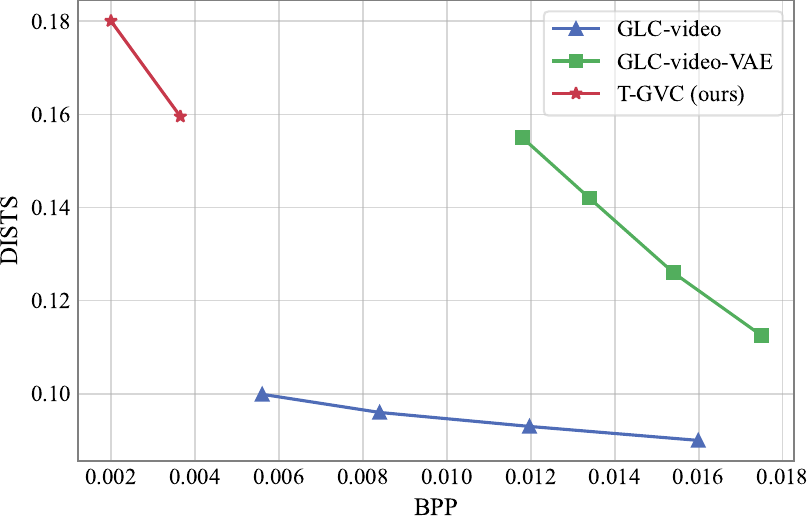}
    \caption{Comparison of the R-D performance between GLC and T-GVC on the UVG dataset.}
    \label{fig8}
\end{figure}

\section{Discussion}

We further compare our method with the recent VQ-VAE-based generative approach GLC-video \cite{qi2025generative} on the UVG dataset (resized to 720p for fair comparison), using DISTS \cite{ding2020dists} as the evaluation metric. The results are shown in the Figure \ref{fig8}, where the performance of GLC-video is reported as in the original paper. Benefiting from a more compact representation of inter-frame dependencies, T-GVC achieves lower bitrate points compared to GLC-video. However, although the highest bitrate point of T-GVC ($\approx$ 0.004 bpp) delivers comparable perceptual quality to the lowest point of GLC-video-VAE ($\approx$0.012 bpp), there remains a noticeable gap between T-GVC and GLC-video. Since GLC-video employs a VQ-VAE-based image encoder that has been shown to outperform MS-ILLM \cite{muckley2023improving} used in T-GVC under ULB conditions, there is potential for T-GVC to achieve superior reconstruction quality at even lower bitrates through further optimization of its image encoder and the design of more refined guidance strategies.

\end{document}